%% file: arXiv.tex
\documentclass[journal]{IEEEtran}
\ifCLASSINFOpdf

\else

\fi
\usepackage{amsmath,amsfonts}
\usepackage{algorithmic}
\usepackage{algorithm}
\usepackage{array}
\usepackage[caption=false,font=normalsize,labelfont=sf,textfont=sf]{subfig}
\usepackage{textcomp}
\usepackage{stfloats}
\usepackage{url}
\usepackage{verbatim}
\usepackage{graphicx}
\usepackage{cite}

\usepackage{multirow} 
\usepackage{booktabs}
\usepackage{hyperref}  

\usepackage{microtype}

\begin{document}



\title{VGD: Visual Geometry Gaussian Splatting for Feed-Forward Surround-view Driving Reconstruction}



\author{
Junhong Lin, Kangli Wang, Shunzhou Wang, Songlin Fan, Ge Li, and Wei Gao,~\IEEEmembership{Senior Member, IEEE}
\thanks{This work was supported by The Major Key Project of PCL (PCL2024A02), Natural Science Foundation of China (62271013, 62031013), Guangdong Provincial Key Laboratory of Ultra High Definition Immersive Media Technology (2024B1212010006), Guangdong Province Pearl River Talent Program (2021QN020708), Guangdong Basic and Applied Basic Research Foundation (2024A1515010155), Shenzhen Science and Technology Program (JCYJ20240813160202004, JCYJ20230807120808017), Shenzhen Fundamental Research Program (GXWD20201231165807007-20200806163656003). (\textit{Corresponding author: Wei Gao.})}

\thanks{Junhong Lin, Kangli Wang, Shunzhou Wang, Songlin Fan, Ge Li, and Wei Gao are with Guangdong Provincial Key Laboratory of Ultra High Definition Immersive Media Technology, School of Electronic and Computer Engineering, Peking University, Shenzhen 518055, China, and Wei Gao is also with Peng Cheng Laboratory, Shenzhen 518066, China (e-mail: \{jhlin42in, kangliwang, slfan\}@stu.pku.edu.cn, \{shunzhouwang, geli, gaowei262\}@pku.edu.cn).}

\thanks{Our code will be available at: https://github.com/JHLin42in/VGD.}

}


\maketitle

\begin{abstract}
Feed-forward surround-view autonomous driving scene reconstruction offers fast, generalizable inference ability, which faces the core challenge of ensuring generalization while elevating novel view quality. Due to the surround-view with minimal overlap regions, existing methods typically fail to ensure geometric consistency and reconstruction quality for novel views. To tackle this tension, we claim that geometric information must be learned explicitly, and the resulting features should be leveraged to guide the elevating of semantic quality in novel views. In this paper, we introduce \textbf{Visual Gaussian Driving (VGD)}, a novel feed-forward end-to-end learning framework designed to address this challenge. To achieve generalizable geometric estimation, we design a lightweight variant of the VGGT architecture to efficiently distill its geometric priors from the pre-trained VGGT to the geometry branch. Furthermore, we design a Gaussian Head that fuses multi-scale geometry tokens to predict Gaussian parameters for novel view rendering, which shares the same patch backbone as the geometry branch. Finally, we integrate multi-scale features from both geometry and Gaussian head branches to jointly supervise a semantic refinement model, optimizing rendering quality through feature-consistent learning. Experiments on nuScenes demonstrate that our approach significantly outperforms state-of-the-art methods in both objective metrics and subjective quality under various settings, which validates VGD's scalability and high-fidelity surround-view reconstruction.
\end{abstract}


\IEEEpeerreviewmaketitle

\section{Introduction}

\begin{figure}[]
\centering
\includegraphics[width=0.485\textwidth]{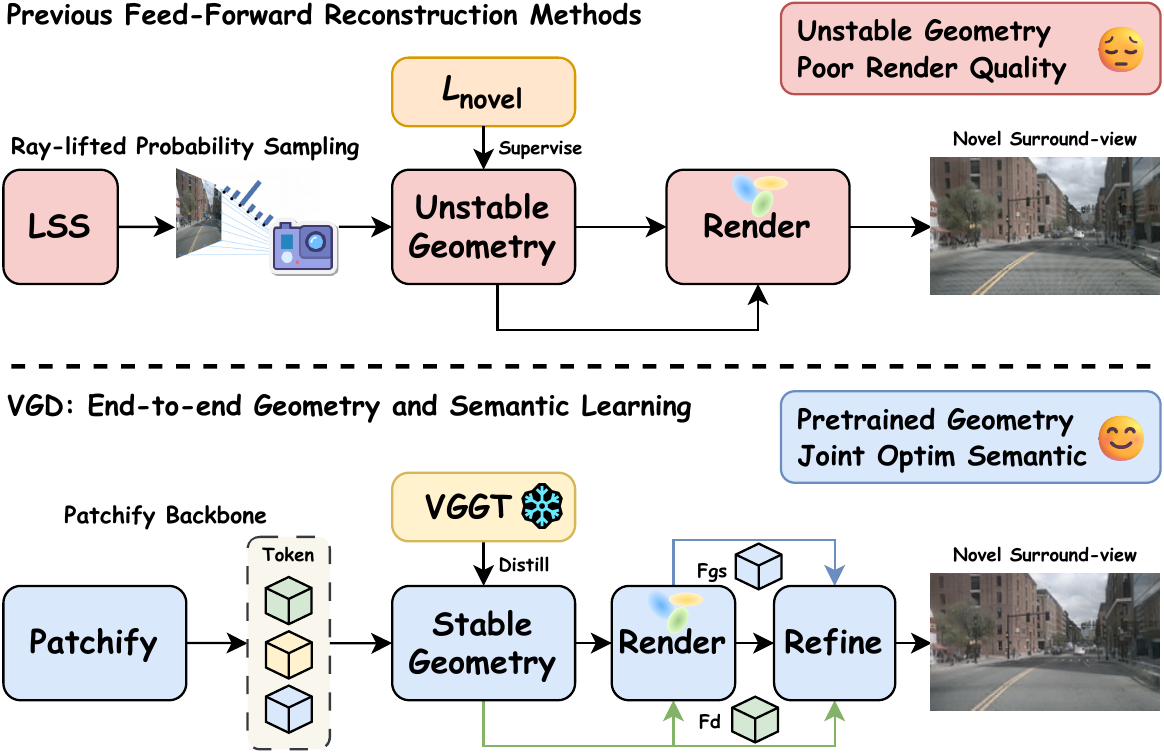}
\caption{Comparison of feed-forward reconstruction paradigms. Previous methods \cite{12_tian2025drivingforward} (top) suffer from poor rendering quality due to unstable 2D-to-3D lifting geometry for direct rendering. In contrast, our VGD (bottom) distills strong geometric priors from a pre-trained model and jointly optimizes semantic information, ensuring geometric stability and high-fidelity rendering.}
\label{fig1-head}
\end{figure}

\IEEEPARstart{S}{parse} surround-view reconstruction is fundamental for understanding autonomous driving scenes and for supplying reliable visual context to downstream modules. Within autonomous driving systems, numerous critical tasks rely on information acquired by multi-camera rigs, including online mapping \cite{1_liao2025maptrv2}, object detection \cite{2-10109207}, occupancy prediction \cite{3_zhang2023occformer}, trajectory prediction \cite{4_hu2023planning}, scene understanding \cite{5_liao2024vlm2scene}, and the construction of high-fidelity simulation environments \cite{6_drivinggaussian}. Accurately recovering scene structure and appearance from sparse, surround-view inputs therefore directly improves the robustness and reliability of these components. In practice, however, surround-view configurations exhibit minimal inter-camera overlap (often below 10\%), wide baselines, heterogeneous intrinsics and extrinsics, and strong viewpoint changes; they also operate under different traffic, frequent occlusions, reflective and textureless surfaces, as well as challenging illumination and weather. These factors jointly degrade geometric cues and increase the risk of cross-view inconsistencies, making high-fidelity 3D reconstruction particularly demanding. The key technical difficulty is to generate geometrically consistent and photometrically faithful novel views from limited observations, while retaining fast, generalizable inference across diverse scenes, camera rigs, and environmental conditions.

The landscape of autonomous driving reconstruction is dominated by two paradigms: optimization-based and feed-forward approaches. General optimization-based methods like NeRF variants \cite{7_mildenhall2021nerf, 48-10375077} offer strong rendering, yet their computational intensity and representation limitations persist. Due to the superior computational efficiency of 3D Gaussian Splatting (3DGS) \cite{8_kerbl20233d, 49-11194205}, optimization-based driving scene methods \cite{6_drivinggaussian, 9_streetgaussian} achieve high-quality novel view synthesis through scene-wide 3DGS modeling at the cost of requiring per-scene optimization. Crucially, these methods lack fast inference capability and cross-scene generalization, limiting autonomous driving applicability. Recent feed-forward models \cite{10_chen2024mvsplat, 11_wang2024dust3r} provide alternatives, yet sparse camera overlap in driving scene undermines their performance. While feed-forward driving frameworks \cite{12_tian2025drivingforward, 13_wei2025omni} have achieved progress, they typically employ LSS (Lift, Splat, Shoot) \cite{33_philion2020lift} or complex schemes to acquire unstable geometric information for direct rendering, generating unsatisfactory results (shown in Figure \ref{fig1-head}).

We argue that a practical reconstruction framework must satisfy three criteria: fast inference capability, generalization, and high reconstruction quality. Fast inference capability demands fast inference speeds, where feed-forward methods demonstrate significant advantages over optimization-based approaches. Generalization requires the model to perform inference in novel scenes post-training, a strength of feed-forward frameworks. Their primary limitation, however, lies in the third criterion. As illustrated in Figure \ref{fig1-head}, minimal inter-camera overlap and large baselines exacerbate depth ambiguity and error accumulation, resulting in failures to produce novel views with sufficient geometric and semantic consistency. Therefore, enhancing sparse novel view reconstruction quality constitutes the primary challenge for advancing feed-forward reconstruction performance.


To bridge this quality gap, we propose a novel feed-forward surround-view reconstruction framework, namely Visual Gaussian Driving (VGD). As shown in Figure \ref{fig1-head}, VGD jointly optimizes visual geometry and semantic representation within a 3D Gaussian Splatting (3DGS) paradigm. The VGD framework comprises three core stages: VGGT Distilled Geometry Prediction, Gaussian Novel View Rendering, and Multi-scale Semantic Refinement. First, to establish a robust geometric foundation, we design a lightweight variant of the VGGT architecture, and distill priors from a pre-trained VGGT model \cite{32_wang2025vggt} into the DPT-Depth branch, ensuring both accuracy and fast inference. Next, guided by the multi-scale geometric features from this branch, we design a Gaussian head (DPT-GS) based on the DPT architecture \cite{32_wang2025vggt} to predict novel view Gaussian parameters, achieving Gaussian novel view rendering. By extracting multi-scale features from the DPT-Depth branch to guide the Gaussian head learning, we ensure geometric consistency in novel view synthesis. Finally, to maximize perceptual quality, a semantic refinement model fuses features from both the geometry and Gaussian pathways for joint optimization.

Our end-to-end VGD framework establishes new state-of-the-art performance in sparse surround-view driving scene reconstruction, achieving significant quality improvements while maintaining fast inference efficiency. Comprehensive nuScenes evaluations quantitatively validate VGD's superiority over both general-purpose and driving-specific methods across multiple settings. Our contributions are summarized as follows:


\begin{itemize}

\item To the best of our knowledge, we propose the first framework that jointly optimizes visual geometry and semantic 3DGS for feed-forward surround-view autonomous driving scene reconstruction. This framework achieves fast, generalizable, and high-quality novel view synthesis.

\item We introduce distilled VGGT for geometric priors and Gaussian prediction. Specifically, we design a lightweight variant of the VGGT model for fast inference with soft distillation for DPT-Depth. To render consistent novel views, we propose Gaussian head DPT-GS that leverages multi-scale geometric features from DPT-Depth.

\item We design a semantic refinement model that hierarchically integrates multi-scale features from DPT-Depth and DPT-GS. This enables joint optimization to elevate novel view synthetic quality.

\item Experimental results demonstrate that VGD achieves state-of-the-art performance, significantly surpassing previous feed-forward autonomous driving reconstruction algorithms under various settings.

\end{itemize}

\section{Related Work}

\subsection{Optimization-based 3D Scene Representation}

Optimization-based methods, which learn a per-scene representation, have demonstrated remarkable capabilities in high-fidelity 3D reconstruction. Neural Radiance Fields (NeRF) \cite{7_mildenhall2021nerf, 48-10375077} and 3D Gaussian Splatting (3DGS) \cite{8_kerbl20233d, 16_studentgs, 50-wu2024deferredgs} have become foundational techniques for novel view synthesis. Due to the extremely fast rendering speed of 3DGS, it has spurred a wave of subsequent research. Recent advancements include Mip-Splatting \cite{14_yu2024mip}, which addresses aliasing and improves stability at high resolutions. DropoutGS \cite{15_xu2025dropoutgs} for improved sparse-view robustness via stochastic regularization. Complementary progress refines shading and editability: GaussianShader \cite{35-Jiang_2024_CVPR} augments splats with shading functions for complex reflectance.


Beyond Gaussians, explicit and hybrid NeRF derivatives \cite{35-fridovich2022plenoxels, 36-yu2021plenoctrees,37-muller2022instant, 38-chen2022tensorf,39-fridovich2023k,40-barron2023zip} have reduced the computational burden of per-scene optimization. Instant-NGP \cite{37-muller2022instant} speeds training dramatically with multiresolution hash grids. Zip-NeRF \cite{40-barron2023zip} brings anti-aliasing to grid-based radiance fields. While these works collectively improve efficiency and quality, the optimization remains per-scene, leading to long fitting times and limited scalability when scenes change rapidly or when thousands of scenes must be processed online.

For autonomous driving, several works \cite{9_streetgaussian, 41-cao2024lightning} have extended NeRF and 3DGS to reconstruct large-scale autonomous driving scene, focusing on detailed 3D or 4D per-scene reconstruction. Street Gaussians \cite{9_streetgaussian} introduces semantically-labeled 3D Gaussians with optimizable poses and dynamic appearance modeling, achieving fast rendering and SOTA performance on autonomous driving benchmarks. However, a significant drawback of these methods is their need for per-scene training, which imposes a prohibitive computational burden and limits their scalability for large-scale applications. In contrast, we aim to develop a feed-forward model with fast inference and generalization capability.

\subsection{Feed-forward 3D Scene Representation}

To overcome the limitations of per-scene optimization, feed-forward methods learn generalizable priors from large-scale datasets. These methods directly infer a scene representation from a set of input views, making them markedly faster at inference than their optimization-based counterparts. Early generalizable NeRFs include pixelNeRF \cite{42-yu2021pixelnerf}, IBRNet \cite{43-wang2021ibrnet}, and MVSNeRF \cite{44-chen2021mvsnerf}, which explore conditioning, image-based rendering, and cost-volume reasoning to produce radiance fields without per-scene fitting. Grid- and anti-aliasing-aware methods such as Zip-NeRF \cite{40-barron2023zip} and extensions to unbounded scenes (e.g., Mip-NeRF 360 \cite{45-barron2022mip}) further improve the quality/speed trade-off.

With 3DGS, several feed-forward models predict Gaussian primitives directly from images. pixelSplat \cite{18_charatan2024pixelsplat} uses pixel-aligned features for realtime, memory-efficient inference. MVSplat \cite{10_chen2024mvsplat} couples a cost-volume encoder with splatting to robustly lift multiview inputs. LGM \cite{19_tang2024lgm} scales multi-view Gaussian generation to high-resolution content. Geometry-centric vision backbones also provide strong priors: DUSt3R \cite{11_wang2024dust3r} regresses dense pointmaps from unposed pairs, MASt3R \cite{46-leroy2024grounding} grounds dense matching in 3D, and VGGT \cite{32_wang2025vggt} jointly predicts cameras, depth, pointmaps, and tracks in a single forward pass—offering generalizable geometric signals.

Addressing the unique challenges of autonomous driving, DistillNeRF \cite{20_wang2024distillnerf} proposed a generalizable model for driving scene that achieves reasonable novel view synthesis, but requires more training resources and additional LiDAR data. More recently, several methods have successfully developed generalizable 3DGS models for this domain, etc. VDG \cite{21_li2025vdg} and Omni-Scene \cite{13_wei2025omni}. DrivingForward \cite{12_tian2025drivingforward} employs self-supervised learning for depth prediction and subsequently utilizes 3DGS for novel view synthesis. Different from them, we aim to further investigate quality constraints for novel views in feed-forward autonomous driving reconstruction.

\begin{figure*}[h]
    \centering
	\includegraphics[width=1\textwidth]{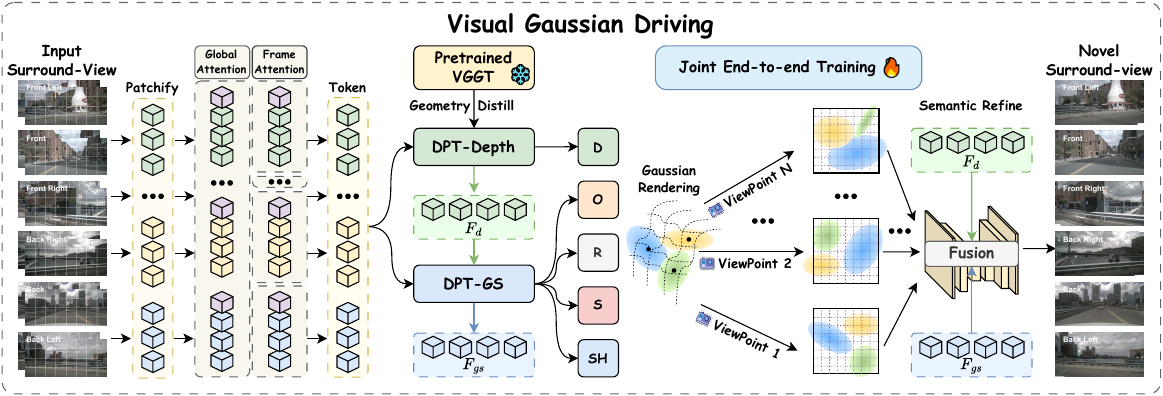}
	\caption{The end-to-end architecture of VGD. Our framework transforms surround-view inputs into novel views in three stages. First, a DPT-Depth branch, supervised by a pre-trained VGGT, extracts strong geometric priors. Second, a DPT-GS head uses these geometric features to guide the prediction of other Gaussian attributes. Finally, a fusion model refines the initial renderings using multi-scale features from both branches to produce the high-fidelity output.}
	\label{fig1-net}
\end{figure*}

\section{Methodology}

\subsection{Problem Definition}

Sparse surround-view reconstruction in autonomous driving is highly challenging due to large baselines, minimal inter-camera overlap (typically $<10\%$), and strong viewpoint variations. Optimization-based methods require costly per-scene optimization and lack generalization, while feed-forward models enable real-time inference yet suffer from unstable geometry and inconsistent novel-view synthesis under sparse inputs. Existing driving-scene feed-forward frameworks often enforce geometric consistency through implicit depth learning, which stabilizes projections but compromises semantic fidelity. Therefore, our goal is to balance speed, generalization, and reconstruction quality by explicitly learning geometry to guide rendering and using refinement to elevate semantic fidelity.


We consider that the primary challenge for feed-forward models lies in simultaneously achieving strong cross-scene generalization while preserving high-fidelity reconstruction. 
Given $N$ input images captured by a calibrated multi-camera rig $\mathcal{I}_{\text{in}}=\{I_i\}_{i=1}^{N}$,
with known camera intrinsics $\mathcal{K}=\{K_i\}_{i=1}^{N}$ and extrinsics 
$\mathcal{T}=\{T_i\}_{i=1}^{N}$, our objective is to synthesize novel views $\mathcal{I}_{\text{nov}}=\{I_m\}_{m=1}^{M}$
corresponding to new target poses 
\(
\mathcal{T}_{\text{nov}}=\{T_m^{\star}\}_{m=1}^{M}
\)
and intrinsics 
\(
\mathcal{K}_{\text{nov}}=\{K_m^{\star}\}_{m=1}^{M}.
\)
Formally, we define the reconstruction process as:
\begin{equation}
\mathcal{I}_{\text{nov}} = \mathcal{R}(\mathcal{I}_{\text{in}}, \mathcal{K}, \mathcal{T}, \mathcal{K}_{\text{nov}}, \mathcal{T}_{\text{nov}}),
\end{equation}
where $\mathcal{R}(\cdot)$ is a feed-forward network with fixed parameters at test time (no scene-specific optimization). 

To address this fundamental limitation, we follow our goal to decompose the reconstruction method $\mathcal{R}(\cdot)$ into a three-stage, geometry-aware pipeline. First, a geometry network $\mathcal{G}$ acquires an explicit geometry representation $\mathcal{O_{\text{geo}}}$ from the input views, which is essential for projecting features to novel viewpoints. Second, we use 3DGS based $\mathcal{S}$ to render $I_{\text{render}}$. Finally, we use $\mathcal{E}$ to get refined $I_{\text{nov}}$. Hereby, we decompose $\mathcal{R}$ into three main modules:
\[
\mathcal{R} = \mathcal{G} \circ \mathcal{S} \circ \mathcal{E},
\]
where $\mathcal{G}$ denotes geometry extraction, $\mathcal{S}$ denotes view synthesis, and $\mathcal{E}$ denotes refinement.

\subsection{Three-Stage Reconstruction Framework}

\paragraph{Geometry Extraction}
The geometry network $\mathcal{G}(\cdot)$ encodes input multi-view images into an explicit geometric representation:
\begin{equation}
\mathcal{O}_{\text{geo}} = \mathcal{G}(\mathcal{I}_{\text{in}}; \theta_G),
\end{equation}
where $\mathcal{O}_{\text{geo}}$ represents a geometry prior. This explicit geometry enables consistent projection of visual features into target viewpoints.

\paragraph{View Synthesis}
Conditioned on $\mathcal{O}_{\text{geo}}$, the synthesis network $\mathcal{S}(\cdot)$ aggregates features across views and predicts a coarse novel-view rendering:
\begin{equation}
I_{\text{render}} = \mathcal{S}\!\left(\mathcal{O}_{\text{geo}}, \mathcal{K}, \mathcal{T}, K_{nov}, T_{nov}, \mathcal{O}_{\text{param}}; \theta_S\right),
\end{equation}
where $\mathcal{O}_{\text{param}}$ are the rendering parameters such as Gaussian attributes (opacity, rotation, scale, and spherical harmonics). 
This stage ensures geometric consistency across large baselines while maintaining computational efficiency.

\paragraph{Semantic Refinement}
Finally, a refinement network $\mathcal{E}(\cdot)$ elevates visual fidelity by leveraging residual learning over the initial rendering:
\begin{equation}
I_{\text{nov}} = \mathcal{E}\!\left(I_{\text{render}}, \mathbf{F}; \theta_E\right),
\end{equation}
where $\mathbf{F}$ denotes multi-scale features extracted from $\mathcal{G}$ and $\mathcal{S}$. 
This step restores high-frequency details and improves semantic coherence, addressing texture degradation typical in sparse-view synthesis.



\subsection{Optimization Goal}

Instead of training these components independently, joint optimization allows the geometry branch to provide reliable 3D structural priors for rendering, while feedback from synthesis and refinement stages promotes geometry representations that are not only geometrically accurate but also perceptually consistent. This collaborative training strategy enables the entire system to learn feature alignment across stages, leading to coherent and high-fidelity reconstruction even under sparse input conditions.

Formally, given ground-truth views $\{I_{gt}\}$, we jointly optimize the parameters of the geometry, synthesis, and refinement networks $(\theta_G, \theta_S, \theta_E)$ by minimizing the reconstruction loss:
\begin{equation}
\min_{\theta_G, \theta_S, \theta_E} 
\frac{1}{M}\sum_{m=1}^{M}
\mathcal{L}\!\left(I_{\text{nov}}, I_{gt}\right).
\end{equation}
Moreover, we follow \cite{12_tian2025drivingforward} to compute three photometric losses $\mathcal{L}_{tm}$, $\mathcal{L}_{sp}$, and $\mathcal{L}_{sp-tm}$ for each camera and smoothness loss \cite{47-godard2017unsupervised} $\mathcal{L}_{smooth}$ for scale-aware localization:
\begin{equation}
\mathcal{L}_{loc} = \mathcal{L}_{tm} + \lambda_{sp}\mathcal{L}_{sp} + \lambda_{sp-tm}\mathcal{L}_{sp-tm} + \lambda_{smooth}\mathcal{L}_{smooth}.
\label{eq:loc_loss}
\end{equation}

This end-to-end training paradigm thus enables efficient and generalizable reconstruction, producing geometrically consistent and visually faithful novel-view synthesis under sparse surround-view conditions.


\section{Network Design}

\subsection{Overall}

To address these challenges, our core insight is to leverage knowledge distilled from pre-trained foundation models to ensure geometric accuracy, while employing architectural similarity-preserving compression to guarantee fast inference and generalization capabilities. We utilize 3D Gaussian Splatting for novel view rendering and guide semantic quality elevation by integrating both geometric and Gaussian-based knowledge. The entire framework undergoes joint end-to-end learning.

As illustrated in Fig. \ref{fig1-net}, we propose \textbf{Visual Gaussian Driving (VGD)}, a novel feed-forward autonomous driving reconstruction framework that jointly optimizes geometric and semantic within a 3D Gaussian Splatting (3DGS) paradigm. First, input images are split into patches $\mathbf{f_{\text{patch}}}$, which undergo multi-layer processing via global and frame attention to produce encoded feature tokens $\mathbf{f_{\text{token}}}$. These tokens are then fed into the DPT-Depth branch $\mathcal{G}(\cdot)$ to generate depth predictions, where we simultaneously apply soft distillation supervision from a pre-trained VGGT model to learn powerful geometric priors $\mathcal{O}_{\text{geo}}$. Next, multi-scale tokens $\mathbf{f}_{\text{d}}$ from the DPT-Depth branch are input to the DPT-GS branch, which predicts Gaussian parameters $\mathcal{O}_{\text{param}}$: opacity ($\mathbf{O}$), rotation ($\mathbf{R}$), scale ($\mathbf{S}$), and spherical harmonics ($\mathbf{SH}$). These parameters, along with transformed coordinates $\mathbf{D}$, enable the 3DGS rasterizer $\mathcal{S}(\cdot)$ to render an initial view, $\mathcal{I_{\text{render}}}$. Finally, the rendered novel view, together with multi-scale features from DPT-Depth $\mathbf{f}_{\text{d}}$ and DPT-GS $\mathbf{f}_{\text{gs}}$ branches, is fed into the semantic refinement model $\mathcal{E}(\cdot)$. Through deep feature fusion, this network produces high-quality novel view outputs $\mathcal{I_{\text{novel}}}$.

VGD establishes a pioneering feed-forward reconstruction architecture with three core advantages: 1) Joint end-to-end geometric-semantic optimization, 2) Efficient knowledge transfer from foundation models and 3) Scalable VGGT-inspired architecture. 


\subsection{VGGT Distilled Geometry Prediction}

Acquiring accurate geometric information from surround-view in autonomous driving scene is a significant challenge. Without effective supervision, models often struggle to learn meaningful geometric constraints, leading to inaccuracies that compromise object consistency across views. To address this, we propose to distill knowledge from a powerful, pre-trained foundation model, specifically adopting the architecture and geometric priors of VGGT as guidance for a more compact and efficient network.

To meet the fast inference and generalization requirements of our feed-forward setting, we design a lightweight variant of the VGGT architecture to efficiently capture its geometric knowledge $\hat{\mathcal{O}}_{\text{geo}}$. Our implementation employs a patch-based paradigm, dividing input views into uniform patches $\mathbf{f_{\text{patch}}}$ and augmenting them with camera tokens. These are processed by a series of global attention $\mathcal{A}_g$ and frame attention $\mathcal{A}_f$ mechanisms to learn cross-patch relationships. The final token representation, $\mathbf{f_{\text{token}}}$, is thus formulated as:
\begin{equation}
\mathbf{f_{\text{token}}} = \left( \mathcal{A}_f \circ \mathcal{A}_g \right)^N \left( \mathbf{f_{\text{patch}}} \right),
\end{equation}
where the composite attention block is applied $N$ times. While maintaining structural similarity to VGGT, our network retains only the DPT-Depth branch, $\mathcal{G}(\cdot)$, but with reduced feature dimensionality for efficiency. This network processes $\mathbf{f_{\text{token}}}$ to produce an initial geometric representation, which we formulate as a disparity map $\mathcal{O_{\text{dis}}}$. This is then converted to metric depth via $\mathcal{O_{\text{geo}}} = \frac{f \cdot B}{\mathcal{O_{\text{dis}}}}$, where $f$ is the focal length and $B$ is the stereo baseline. This step explicitly incorporates camera geometry, ensuring consistency across novel viewpoints and diverse sensor configurations.

Geometric knowledge is transferred via soft distillation, where we supervise our model's depth output, $\mathcal{O}_{\text{geo}}$, against the predictions from the pre-trained VGGT teacher, $\hat{\mathcal{O}}_{\text{geo}}$. 
We employ a dual-loss mechanism $\mathcal{L}_{\text{distill}}$ that combines a Smooth L1 loss \cite{25_girshick2015fast} and affine-invariant loss \cite{26_ranftl2020towards}:
\begin{equation}
\begin{aligned}
\mathcal{L}_{\text{smooth-L1}} &= \sum_i 
\begin{cases} 
0.5 (\delta)^2,  & |\delta|<1, \\ 
|\delta|-0.5, & \text{otherwise},
\end{cases} \\
\mathcal{L}_{\text{affine}} &= \left\| \log\frac{\mathcal{O}_{\text{geo}}}{\text{med}(\mathcal{O}_{\text{geo}})} - \log\frac{\hat{\mathcal{O}}_{\text{geo}}}{\text{med}(\hat{\mathcal{O}}_{\text{geo}})} \right\|_2,
\end{aligned}
\end{equation}
where $\delta = \mathcal{O}_{\text{geo\_i}} - \hat{\mathcal{O}}_{\text{geo\_i}}$. During this distillation, we also extract multi-scale features $\mathbf{f}_{\text{d}} = \{f_{d}^{(k)}\}_{k=1}^4$ from the DPT-Depth branch to provide hierarchical geometric context for subsequent processing stages.

The resulting distilled model achieves a 95\% parameter reduction compared to the original VGGT while maintaining fast inference and robust generalization. With these accurate geometric priors and multi-scale features ($\mathcal{O_{\text{geo}}}$ and $\mathbf{f}_{\text{d}}$) now established, the next critical step is to leverage this structural foundation to render the visual appearance of novel views.

\subsection{Gaussian Novel View Rendering}

While 3DGS is a powerful rendering technique, its reliance on dense multi-view optimization fundamentally limits its application in sparse surround-view autonomous driving scene. To overcome this limitation, we leverage the high-quality geometric priors established in the previous step to introduce a joint prediction framework that accurately predicts Gaussian parameters for sparse surround-view.

Hereby, we design a Gaussian prediction head (DPT-GS) based on the DPT architecture to predict Gaussian parameters $\mathcal{O}_{\text{param}}$. This head takes as input the multi-scale features $\mathbf{f}_{\text{d}}$ extracted from the DPT-Depth branch, $\mathcal{G}(\cdot)$. By fusing these geometric features into the DPT-GS branch, we explicitly constrain the prediction of Gaussian parameters, ensuring they maintain spatial coherence with the underlying scene structure. The shared tokens $\mathbf{f_{\text{token}}}$ from the patch-based backbone are also fed into DPT-GS, enabling joint end-to-end optimization. The DPT-GS head $\phi(\cdot)$ predicts all Gaussian parameters through specialized convolutional pathways:

\begin{equation}
\begin{aligned}
\mathbf{f}_g &= \phi(\mathbf{f}_{\text{d}}, \mathbf{f_{\text{token}}}), \\ 
\mathbf{R} &= \sigma(\mathbf{W_r} \ast \mathbf{f}_g), \\
\mathbf{O} &= \text{sigmoid}(\mathbf{W_\alpha} \ast \mathbf{f}_g), \\
\mathbf{S} &= \exp(\mathbf{W_s} \ast \mathbf{f}_g), \\
\mathbf{SH} &= \mathbf{W_{sh}} \ast \mathbf{f}_g,
\end{aligned}
\end{equation}
where $\mathbf{f}_g$ represents the final feature map within the DPT-GS branch, and $\sigma(\cdot)$ normalizes the output to valid quaternions for rotation $\mathbf{R}$.
Concurrently, $\mathcal{O_{\text{geo}}}$ are transformed into 3D Gaussian mean parameters $\mathbf{D}$ via camera coordinate conversion:
\begin{equation}
\mathbf{D} = \pi^{-1}(\mathcal{O_{\text{geo}}}, \mathbf{K}, \mathbf{T}),
\end{equation}
where $\mathbf{K}$ and is the intrinsic matrix and $\mathbf{T}$ the extrinsic transformation. Multi-scale features $\mathbf{f}_{\text{gs}} = \{f_{gs}^{(k)}\}_{k=1}^4$ are extracted from DPT-GS for subsequent refinement stages. The complete set of Gaussians, $\mathcal{G} = \{\mathbf{D}_i, \mathbf{R}_i, \mathbf{O}_i, \mathbf{S}_i, \mathbf{SH}_i\}_{i=1}^N$, is then rendered into the initial novel view $\mathcal{I_{\text{render}}}$ via a differentiable rasterizer:
\begin{equation}
    \mathcal{I_{\text{render}}} = \sum_{i \in \text{sorted}} \mathbf{C}_i \alpha_i \prod_{j=1}^{i-1} (1-\alpha_j),
\end{equation}
where $\mathbf{C}_i$ and $\alpha_i$ are the color and alpha values of the $i$-th Gaussian, computed from its parameters ($\mathbf{SH}_i$, $\mathbf{O}_i$) and projected covariance.

This integrated approach produces an initial rendered view, $\mathcal{I_{\text{render}}}$, with strong geometric consistency. However, direct feed-forward synthesis can still introduce subtle artifacts or a lack of fine detail. Therefore, a final refinement stage is necessary to elevate the rendering fidelity.




\subsection{Multi-scale Semantic Refinement}

Initial renderings from a feed-forward 3DGS, especially in surround-view autonomous driving scene, often suffer from artifacts and quality degradation. To address this, as motivated in the previous section, we propose a Semantic Refinement Model (SRM) that enhances the final image quality by conditioning on both geometric and appearance-based features from our dual-branch backbone.

The SRM takes as input the initial rendered view, $\mathcal{I_{\text{render}}}$, and a deep fusion of multi-scale features extracted from both the DPT-Depth branch $\mathbf{f}_{\text{d}}$ and the DPT-GS branch $\mathbf{f}_{\text{gs}}$. This fusion allows for a synergistic interplay between geometric coherence from $\mathbf{f}_{\text{d}}$ and semantic consistency from $\mathbf{f}_{\text{gs}}$, enabling the model to learn robust, viewpoint-invariant representations. The final, high-quality reconstruction $\mathcal{I_{\text{novel}}}$ is then obtained via residual learning:
\begin{equation}
\mathcal{I_{\text{novel}}} = \mathcal{I_{\text{render}}} + \mathcal{E}(I_{\text{render}}, \mathbf{f}_{\text{d}} \oplus \mathbf{f}_{gs}),
\end{equation}
where $\mathcal{E}(\cdot)$ is the SRM and $\oplus$ denotes feature concatenation. This residual formulation is critical, as it constrains the network to learn only the necessary quality improvements, which stabilizes training and mitigates the over-smoothing artifacts common in full image regeneration tasks.

After obtaining $\mathcal{I_{\text{novel}}}$, we use loss $\mathcal{L}_{\text{render}}$ to optimize VGD, formulated as a weighted combination of perceptual and pixel-level metrics:
\begin{align}
\mathcal{L}_{\text{render}} &= \lambda_1 \mathcal{L}_{\text{L1}} + \lambda_2 \mathcal{L}_{\text{SSIM}} + \lambda_3 \mathcal{L}_{\text{LPIPS}}.
\end{align}
This multi-faceted loss formulation simultaneously preserves structural integrity through \cite{23_wang2004image} $\mathcal{L}_{\text{SSIM}}$, enforces pixel-level accuracy via $\mathcal{L}_{\text{L1}}$, and maintains perceptual quality with \cite{24_zhang2018unreasonable} $\mathcal{L}_{\text{LPIPS}}$. 

The SRM employs an extremely minimalist U-Net design with less than 0.2\% of VGGT's parameters. This integrated approach establishes an end-to-end model that elevates rendering quality while maintaining fast inference.

\begin{figure*}[]
    \centering
	\includegraphics[width=0.93\textwidth]{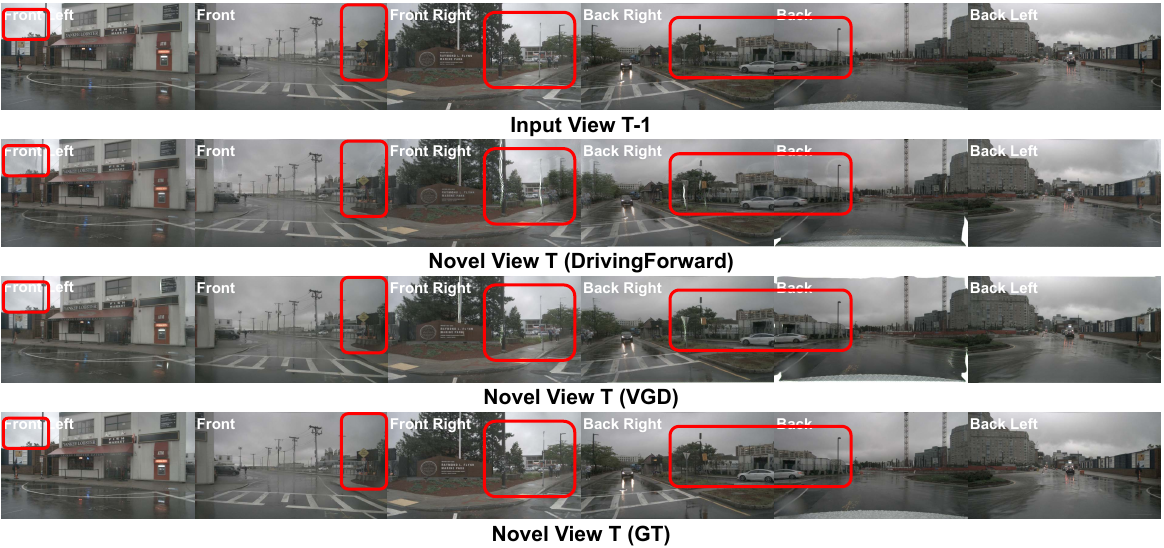}
	\caption{Qualitative results of novel surround-view on Single-Frame Mode. Red rectangular boxes highlight the differences. }
	\label{fig2-sf1}
\end{figure*}

\input{tables/tab}

\subsection{Joint End-to-end Training}

The entire VGD framework is optimized end-to-end to deeply integrate the geometric and semantic pathways, enabling the reconstruction of high-quality novel views from sparse inputs. The training is guided by a composite loss function, $\mathcal{L}$, which comprises three primary components.
The first is the VGGT distillation loss $\mathcal{L}_{\text{distill}}$, which is as detailed in the previous section, transfers robust geometric priors from the pre-trained teacher model to our depth estimation branch. The second component is the rendering loss $\mathcal{L}_{\text{render}}$, which supervises the quality of the final novel view synthesis. The third is scale-aware localization $\mathcal{L}_{\text{loc}}$. The total loss is:
\begin{equation}
\mathcal{L} =  \lambda_r \mathcal{L}_{\text{render}} + \lambda_d \mathcal{L}_{\text{distill}} + \lambda_{l} \mathcal{L}_{\text{loc}},
\end{equation}
where we set $\lambda_{r}$ = 0.01, $\lambda_{d}$ = 0.0005, $\lambda_{l}$ = 1 and train our model for 10 epochs. We use Adam optimizer with the learning rate of $1 \times 10^{-4}$, a batch size of 2 with 6 surround-view images as one sample. We implement VGD with PyTorch of bfloat16, and use a prebuilt 3DGS renderer \cite{8_kerbl20233d}. All experiments are conducted on a single A100 GPU.


\section{Experiments}
\subsection{Evaluation}
\paragraph{Dataset} We utilize the publicly available nuScenes dataset \cite{27_caesar2020nuscenes}, which comprises 1000 diverse driving scene across varied geographic regions. Each scene contains approximately 20 seconds of sequential data, with a total of around 40000 densely annotated keyframes sampled at 2Hz. Image acquisition leverages six vehicle-mounted cameras that provide full surround coverage with constrained inter-camera overlap (approximately 10\%). Following the standard evaluation protocol, we adopt the established dataset partition of 700 training scenes and 150 validation scenes. This comprehensive benchmark enables rigorous assessment of surround-view synthesis performance under authentic autonomous driving conditions.

\paragraph{Metrics} We measure performance through three complementary metrics: peak signal-to-noise ratio (PSNR), structural similarity index (SSIM) \cite{23_wang2004image}, and learned perceptual image patch similarity (LPIPS) \cite{24_zhang2018unreasonable}. These image quality assessment metrics provide a rigorous framework for objectively evaluating the rendering quality of novel views.

\paragraph{Settings} We follow \cite{12_tian2025drivingforward} to conduct comprehensive settings in both single-frame and multi-frame modes, which synthesizes novel views for frames adjacent to input images. We validate the model's performance under sparse surround-view through these two mode settings, demonstrating its generalization and scalability. We validate the model's performance under sparse surround-view through these two mode settings, demonstrating its generalization and scalability.

\subsection{Comparison on Single-Frame Mode}
The Single-Frame (SF) mode aims to synthesize surround-view images at timesteps $t+1$ from surround-view inputs at timesteps $t$.

\paragraph{Compared Methods} We compare our VGD against DrivingForward \cite{12_tian2025drivingforward}, DistillNeRF \cite{20_wang2024distillnerf} and its comparison methods including EmerNeRF \cite{28_yang2023emernerf}, Unipad \cite{29_yang2024unipad}, and SelfOcc \cite{30_huang2024selfocc}. Since DistillNeRF's code is unavailable and unreproducible, we ensure fair comparison by adhering to the experimental specifications detailed in \cite{20_wang2024distillnerf} under the image resolution of $114 \times 228$. To validate high-resolution performance capabilities, we perform a comparative analysis with DrivingForward at the standardized resolution of $352 \times 640$.

\begin{figure*}[]
    \centering
	\includegraphics[width=0.95\textwidth]{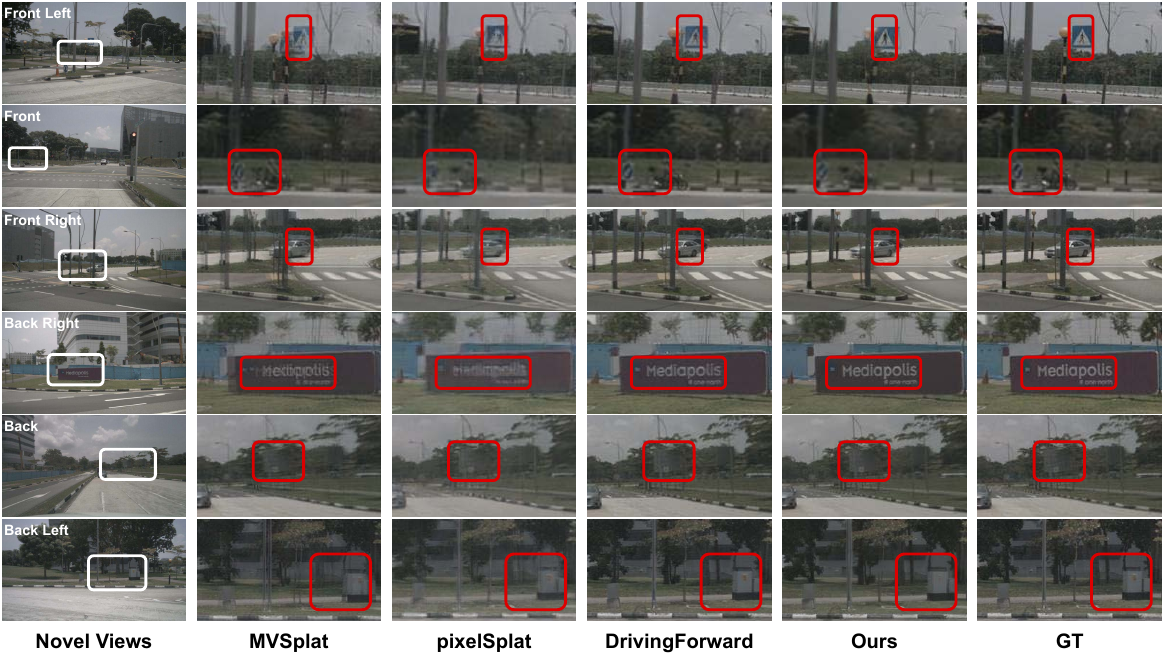}
	\caption{Qualitative results of novel surround-view on Multi-Frame Mode. Red rectangular boxes highlight the differences. }
	\label{fig2-mf}
\end{figure*}

\paragraph{Quantitative Analysis}  Table \ref{tab} presents quantitative results under SF mode across varying resolutions. Our approach demonstrates consistent superiority over all compared methods in every evaluation metric. Notably, it achieves +1.69 dB (high resolution) and +2.11 dB (low resolution) PSNR gains against the feed-forward driving method DrivingForward. These significant improvements validate the resolution robustness and generalization capability of our method.


\paragraph{Qualitative Analysis} Figure \ref{fig2-sf1} demonstrates qualitative results under the single-frame mode. As shown, DrivingForward exhibits noticeable artifacts in reconstructed details, such as distorted utility poles, sky ghosting, and blurred billboards. In contrast, our method achieves reconstruction quality that closely matches the ground truth, demonstrating superior fidelity and visual consistency compared to existing feed-forward reconstruction frameworks. Moreover, the reconstructed scenes from our approach preserve sharper structural edges and more realistic textures, highlighting the effectiveness of our geometry-guided and refinement-based design.

\subsection{Comparison on Multi-Frame Mode}
The Multi-Frame (MF) mode aims to synthesize surround-view images at timesteps $t$ from surround-view inputs at timesteps $t-1$ and $t+1$.

\paragraph{Compared Methods} We compare our method with MVSplat \cite{10_chen2024mvsplat}, pixelSplat \cite{18_charatan2024pixelsplat}, which are designed for training on datasets with densely overlapping inputs. Given that temporally adjacent frames have significantly more overlap than spatially adjacent frames, these methods are suitable for comparison in MF mode. Moreover, we compare DrivingForward \cite{12_tian2025drivingforward} for feed-forward driving scene reconstruction model analysis. All resolutions are set to $352 \times 640$. for feed-forward driving scene reconstruction model analysis.

\paragraph{Quantitative Analysis} As quantified in Table \ref{tab}, our method achieves state-of-the-art performance across all metrics in MF mode. Compared to general-purpose feed-forward models, it demonstrates 2.07 dB$\sim$4.24 dB higher PSNR and 8.9\%$\sim$25.9\% higher SSIM, revealing their inherent limitations in driving scene. Against the driving-specific DrivingForward, VGD shows a +1.01 dB PSNR gain with 1.4\%/1.8\% improvements in SSIM/LPIPS. Furthermore, our inference speed is 3.5$\times$$\sim$7.5$\times$ faster than general models and typically 1.6$\times$ faster than driving-specific methods, validating superior applicability for autonomous driving scene.

\begin{figure*}[]
    \centering
	\includegraphics[width=0.9\textwidth]{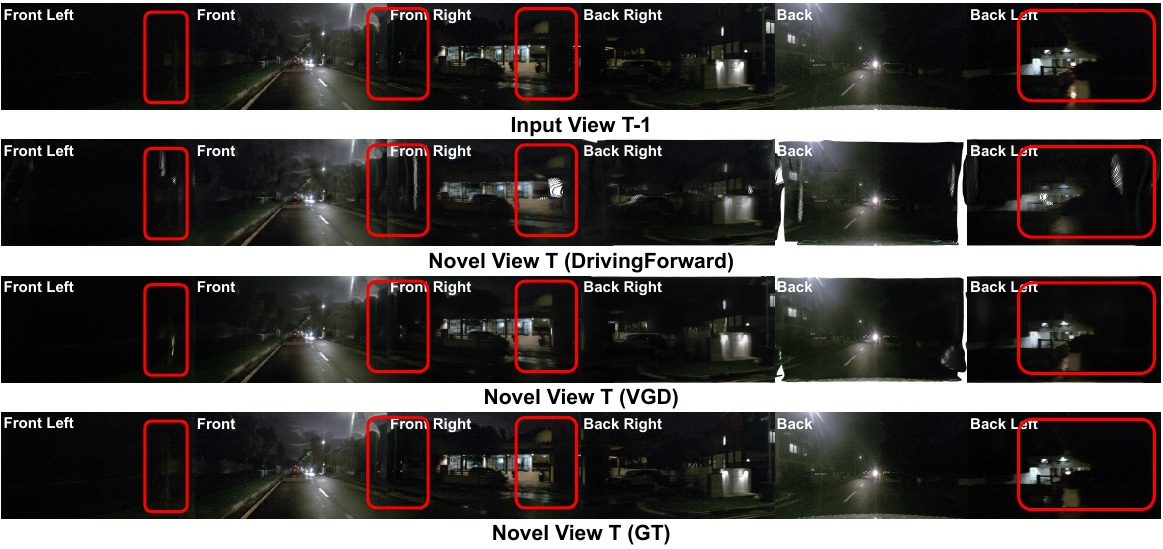}
	\caption{Qualitative results of extreme case on Single-Frame Mode. Red rectangular boxes highlight the differences. }
	\label{fig2-sf2}
\end{figure*}

\begin{figure*}[]
    \centering
	\includegraphics[width=0.9\textwidth]{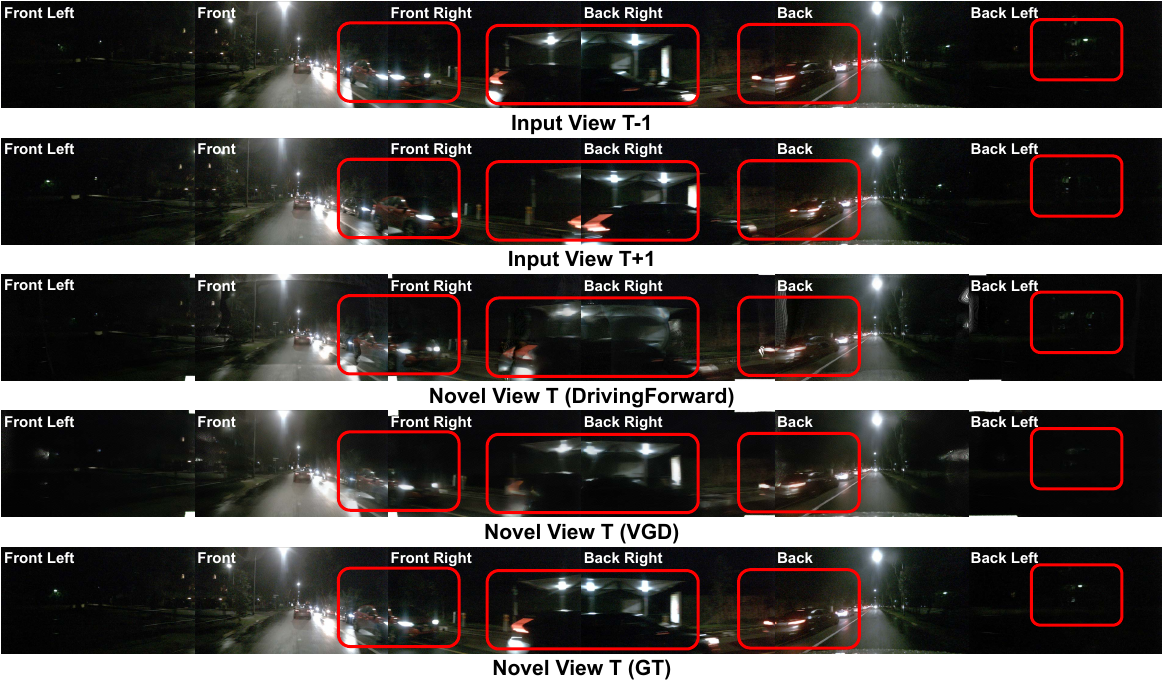}
	\caption{Qualitative results of extreme case on Multi-Frame Mode. Red rectangular boxes highlight the differences. }
	\label{fig2-mf2}
\end{figure*}

\paragraph{Qualitative Analysis} Figure \ref{fig2-mf} details reconstruction visual comparisons. MVSplat and pixelSplat exhibit significant artifact overlap and motion blur in novel views, causing severe geometric inconsistency and semantic information loss. While DrivingForward improves visual quality, it shows persistent misalignment in critical structures (e.g., traffic signs in front-left view, vehicle contours in front/front-right views) and high-frequency texture degradation in semantic regions (e.g., billboard text in back-right view). In contrast, our method achieves superior geometric and semantic consistency, demonstrating the capability to attain higher-fidelity surround-view reconstruction for autonomous driving scene.

\subsection{Comparison on Extreme Cases}

We evaluate reconstruction performance under extreme nighttime conditions in both SF and MF modes, comparing our framework with the previous autonomous driving method DrivingForward, to validate its novelty and robustness. As shown in Figure \ref{fig2-sf2}, DrivingForward fails to reconstruct objects accurately in the SF mode, exhibiting noticeable geometric misalignment in the back-left view, whereas our method maintains stable and precise geometry. Furthermore, our results preserve finer structural details and more realistic textures that closely resemble the ground truth. Figure \ref{fig2-mf2} presents the MF mode results, where DrivingForward suffers from severe quality degradation and inaccurate reconstruction of light sources and vehicles. In contrast, our method effectively preserves geometric consistency and restores realistic lighting effects, achieving visual fidelity comparable to the GT. These results under extreme nighttime scenarios demonstrate that our framework consistently delivers more reliable and high-quality reconstructions than existing feed-forward approaches.


\subsection{Ablation Studies}
We conduct ablation studies to validate the effectiveness of our method. All experiments are trained in SF mode under consistent experimental settings. We perform the following key ablation analyzes of VGD. Geometry Learning \textbf{(GL)} indicates soft distilled learning from pretrained VGGT; Semantic Refine \textbf{(SR)} denotes the use of SRM after Gaussian rendering; Multi-scale \textbf{(MS)} signifies that SRM fuses multi-scale features from both DPT-Depth and DPT-GS; Joint Training \textbf{(JT)} represents joint end-to-end optimization of the entire network. Baseline Setup \textbf{(BS)} denotes a configuration setting without any of these four setups.

\begin{figure*}[]
    \centering
	\includegraphics[width=1\textwidth]{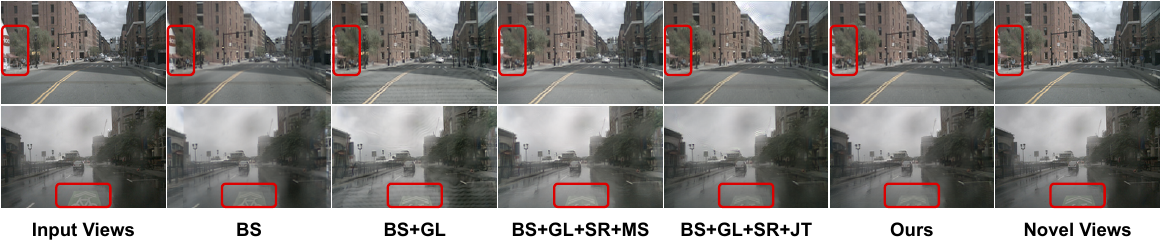}
        \caption{Qualitative results of ablation studies. Red rectangular boxes highlight the differences.}
	\label{fig3-abs}
\end{figure*}

\input{tables/tab3_abs}

\paragraph{The Effect of Geometry Learning} First, we validate the effectiveness of \textbf{GL}. As shown in row 1 of Table \ref{tab3-abs}, \textbf{BS} achieves a PSNR of 22.22 dB and SSIM of 0.715. After incorporating geometric guidance \textbf{(BS+GL)} as presented in row 2, PSNR decreases by 0.7 dB to 21.52 dB while SSIM increases by 0.042 to 0.757. This phenomenon occurs because, without geometric guidance, objects from input views are incorrectly positioned in novel views. As demonstrated by landmarks and buildings in Figure \ref{fig3-abs}, \textbf{BS} outputs maintain near-identical appearance to input images without novel view rendering artifacts, but exhibit structural inconsistencies with target novel views. This results in misleadingly high PSNR but low SSIM.

\paragraph{The Effect of Joint Training Efficacy} To verify joint training efficacy, we first train the Gaussian rendering output, then freeze the front-end model to train the SRM while retaining multi-scale feature fusion (\textbf{BS+GL+SR+MS}, row 3 of Table \ref{tab3-abs}). This configuration shows limited performance gains after adding SRM alone, but improves visual quality in Figure \ref{fig3-abs}. However, our joint end-to-end training approach achieves a 2.14 dB PSNR improvement and more accurate color reproduction, demonstrating that simply adding a refinement module without integrated optimization fails to deliver substantial gains.

\paragraph{The Effect of Multi-scale Fusion} To validate multi-scale fusion effectiveness, we maintain joint end-to-end training (\textbf{BS+GL+SR+JT}, row 4 of Table \ref{tab3-abs}). Performance significantly surpasses the configuration without \textbf{JT}, yet remains 0.84 dB lower than our complete model. This 15.7\% performance gap proves the critical importance of multi-scale feature fusion in joint optimization. Figure \ref{fig3-abs} demonstrates our method's superior structural detail preservation and perceptual quality, achieving the closest resemblance to ground truth. In summary, ablation studies confirm the critical contributions of these four components.

\section{Conclusion}

We propose VGD, a novel feed-forward autonomous driving reconstruction framework that jointly optimizes geometry and semantic representations within a 3D Gaussian Splatting (3DGS) paradigm. We design a lightweight variant VGGT model and distill geometry to achieve fast inference while preserving strong generalization capabilities and geometric consistency. Guided by these strong geometry prior tokens, a Gaussian prediction head DPT-GS is designed to predict Gaussian parameters. Finally, a semantic refinement model fuses multi-scale features from both geometry and Gaussian pathways to elevate visual quality. Comprehensive experiments demonstrate VGD's superior performance, scalability and robustness under sparse surround-view conditions through end-to-end joint optimization. Future work will explore VGD's scalability across diverse frames and datasets in autonomous driving scene. 

{
\small
\bibliographystyle{IEEEtran}
\bibliography{refs}
}

\end{document}

%% file: tables/tab.tex
\begin{table*}[t]
\centering
\caption{Quantitative results on Multi-Frame (MF) and Single-Frame (SF) Modes. Higher is better for PSNR/SSIM; lower is better for LPIPS/Inference. “—” denotes not reported. \textbf{Bold} indicates the best performance.}
\setlength{\tabcolsep}{6pt}
\begin{tabular}{c|c|c|l|cccc}
\toprule
\textbf{Mode} & \textbf{Resolution} & \textbf{Methods} & \textbf{Pub\&Year} & \textbf{PSNR$\uparrow$} & \textbf{SSIM$\uparrow$} & \textbf{LPIPS$\downarrow$} & \textbf{Inference$\downarrow$} \\
\midrule
\multirow{4}{*}{\textbf{Multi Frame}} & \multirow{4}{*}{$352\times640$}
& MVSplat \cite{10_chen2024mvsplat} &ECCV'24            & 22.83 & 0.629 & 0.317 & 1.39 s \\
& & pixelSplat \cite{18_charatan2024pixelsplat} &CVPR'24        & 25.00 & 0.727 & 0.298 & 2.95 s \\
& & DrivingForward \cite{12_tian2025drivingforward} &AAAI'25    & 26.06 & 0.781 & 0.215 & 0.63 s \\
& & \textbf{Ours}   & —             & \textbf{27.07} & \textbf{0.792} & \textbf{0.211} & \textbf{0.39 s} \\
\midrule
\multirow{8}{*}{\textbf{Single Frame}} & \multirow{2}{*}{$352\times640$}
& DrivingForward \cite{12_tian2025drivingforward}&AAAI'25     & 21.67 & 0.727 & 0.259 & 0.35 s \\
& & \textbf{Ours}    & —            & \textbf{23.36} & \textbf{0.749} & \textbf{0.231} & \textbf{0.26 s} \\
\cmidrule{2-8}
& \multirow{6}{*}{$114\times228$}
& UniPad \cite{29_yang2024unipad}&CVPR'24             & 16.45 & 0.375 & —     & — \\
& & SelfOcc \cite{30_huang2024selfocc}&CVPR'24          & 18.22 & 0.464 & —     & — \\
& & EmerNeRF \cite{28_yang2023emernerf}&ICLR'24          & 20.95 & 0.585 & —     & — \\
& & DistillNeRF \cite{20_wang2024distillnerf}& NeurIPS'24    & 20.78 & 0.590 & —     & — \\
& & DrivingForward \cite{12_tian2025drivingforward}&AAAI'25    & 21.76 & 0.767 & 0.194 & — \\
& & \textbf{Ours}   & —             & \textbf{23.87} & \textbf{0.802} & \textbf{0.151} & — \\
\bottomrule
\end{tabular}

\label{tab}
\end{table*}

%% file: tables/tab3_abs.tex


\begin{table}[]
\centering
\caption{Quantitative results of ablation studies.}
\renewcommand{\arraystretch}{1.2}

\begin{tabular}{cccc|ccc}
\toprule
\textbf{GL} & \textbf{SR} & \textbf{MS} & \textbf{JT} & \textbf{PSNR ↑} & \textbf{SSIM ↑} & \textbf{LPIPS ↓} \\ \midrule
$\times$ & $\times$ & $\times$ & $\times$ & 22.22 & 0.715 & 0.197  \\
$\checkmark$ & $\times$ & $\times$ & $\times$ & 21.52 & 0.757 & 0.204 \\
$\checkmark$ & $\checkmark$ & $\checkmark$ & $\times$ & 21.73 & 0.777 & 0.171 \\
$\checkmark$ & $\checkmark$ & $\times$ & $\checkmark$ & 23.03 & 0.787 & 0.169 \\
\midrule
\textbf{$\checkmark$} & \textbf{$\checkmark$} & \textbf{$\checkmark$} & \textbf{$\checkmark$} & \textbf{23.87} & \textbf{0.802} & \textbf{0.151} \\
\bottomrule
\end{tabular}
\label{tab3-abs}
\end{table}